\documentclass[sigconf]{acmart}

\usepackage{booktabs} 
\usepackage{graphicx}
\usepackage{epsfig}
\usepackage{float}
\usepackage{array}
\restylefloat{table}
\usepackage{ragged2e}
\usepackage{caption}
\usepackage{tabularx}
\usepackage{amssymb,amsmath,mathrsfs}
\usepackage{graphicx}
\usepackage[linesnumbered,ruled,lined]{algorithm2e}
\usepackage[font=small,labelfont=bf]{caption}
\usepackage[small]{subfigure}
\usepackage{color}
\usepackage{url}
\usepackage{float}
\usepackage{multirow}
\usepackage{tabularx}
\usepackage{bbm}
\usepackage{enumitem}

\newtheorem{hypothesis}{Hypothesis}

\newcommand{\nop}[1]{}

\newcolumntype{P}[1]{>{\RaggedRight\hspace{0pt}}p{#1}}
\newcolumntype{L}[1]{>{\raggedright\let\newline\\\arraybackslash\hspace{0pt}}m{#1}}
\newcolumntype{C}[1]{>{\centering\let\newline\\\arraybackslash\hspace{0pt}}m{#1}}
\newcolumntype{R}[1]{>{\raggedleft\let\newline\\\arraybackslash\hspace{0pt}}m{#1}}

\newcommand{\secref}[1]{Section \ref{#1}}
\newcommand{\figref}[1]{Figure \ref{#1}}

\newcommand{\tabref}[1]{Table \ref{#1}}

\newcommand{\cut}[1]{}

\DeclareMathOperator{\RR}{\mathbb{R}}
\DeclareMathOperator{\EE}{\mathbb{E}}

\def\e{\mathrm{exp}}
\def\log{\mathrm{log}}

\def\ie{{\sl i.e.}}
\def\eg{{\sl e.g.}}

\def\L{\mathcal{L}}
\def\O{\mathcal{O}}
\def\F{\mathcal{F}}

\DeclareMathOperator{\R}{\mathcal{R}}
\DeclareMathOperator{\Rn}{\overline{\mathcal{R}}}
\def\I{\mathcal{I}}
\def\Z{\mathcal{Z}}
\def\M{\mathcal{M}}
\def\Y{\mathcal{Y}}

\def\D{\mathcal{D}}
\def\Q{\mathcal{Q}}
\def\A{\mathcal{A}}
\def\P{\mathcal{P}}

\def\E{\mathcal{E}}
\def\T{\mathcal{T}}

\def\bc{\mathbf{c}}

\def\bz{\mathbf{z}}
\def\br{\mathbf{r}}

\def\bp{\mathbf{p}}

\newcolumntype{x}{>{\hsize=.8\hsize}X}
\newcolumntype{a}{>{\hsize=.4\hsize}X}
\newcolumntype{b}{>{\hsize=1.6\hsize}X}
\setcopyright{acmcopyright}

\begin{document}

\acmDOI{10.1145/3159652.3159709}

\acmISBN{978-1-4503-5581-0/18/02}

\copyrightyear{2018}
\acmYear{2018}
\setcopyright{acmcopyright}
\acmConference[WSDM 2018]{WSDM 2018: The Eleventh ACM International Conference on Web Search and Data Mining }{February 5--9, 2018}{Marina Del Rey, CA, USA}
%
\acmPrice{15.00}

\title{Indirect Supervision for Relation Extraction using Question-Answer Pairs}
%

\author{Zeqiu Wu}
\affiliation{
	\institution{University of Illinois, Urbana-Champaign}
}
\email{zeqiuwu1@illinois.edu}

\author{Xiang Ren}
\affiliation{
	\institution{University of Southern California}
    \city{Los Angeles}
    \state{California}
}
\email{xiangren@usc.edu}

\author{Frank F. Xu}
\affiliation{
	\institution{Shanghai Jiao Tong University}
    \city{Shanghai}
    \country{China}
}
\email{frankxu@sjtu.edu.cn}

\author{Ji Li}
\affiliation{
	\institution{University of Illinois, Urbana-Champaign}
}
\email{jili3@illinois.edu}

\author{Jiawei Han}
\affiliation{
	\institution{University of Illinois, Urbana-Champaign}
}
\email{hanj@illinois.edu}

\renewcommand{\shortauthors}{Z. Wu et al.}

\begin{abstract}
Automatic relation extraction (RE) for types of interest is of great importance for interpreting massive text corpora in an efficient manner. For example, we want to identify the relationship ``\texttt{president\_of}'' between entities ``\textit{Donald Trump}'' and ``\textit{United States}'' in a sentence expressing such a relation. Traditional RE models have heavily relied on human-annotated corpus for training, which can be costly in generating labeled data and become obstacles when dealing with more relation types. Thus, more RE extraction systems have shifted to be built upon training data automatically acquired by linking to knowledge bases (distant supervision). However, due to the incompleteness of knowledge bases and the context-agnostic labeling, the training data collected via distant supervision (DS) can be very noisy. In recent years, as increasing attention has been brought to tackling question-answering (QA) tasks, user feedback or datasets of such tasks become more accessible. In this paper, we propose a novel framework, \textsc{ReQuest}, to leverage question-answer pairs as an indirect source of supervision for relation extraction, and study how to use such supervision to reduce noise induced from DS. Our model jointly embeds relation mentions, types, QA entity mention pairs and text features in two low-dimensional spaces (RE and QA), where objects with same relation types or semantically similar question-answer pairs have similar representations. Shared features connect these two spaces, carrying clearer semantic knowledge from both sources. \textsc{ReQuest}, then use these learned embeddings to estimate the types of test relation mentions. We formulate a global objective function and adopt a novel margin-based QA loss to reduce noise in DS by exploiting semantic evidence from the QA dataset. Our experimental results achieve an average of 11\% improvement in F1 score on two public RE datasets combined with TREC QA dataset. Codes and datasets can be downloaded at \url{https://github.com/ellenmellon/ReQuest}.
\end{abstract}

%
%
%
%
%
%

\maketitle

\begin{figure}[t]
	\centering
	\begin{small}
    \includegraphics[width=\columnwidth]{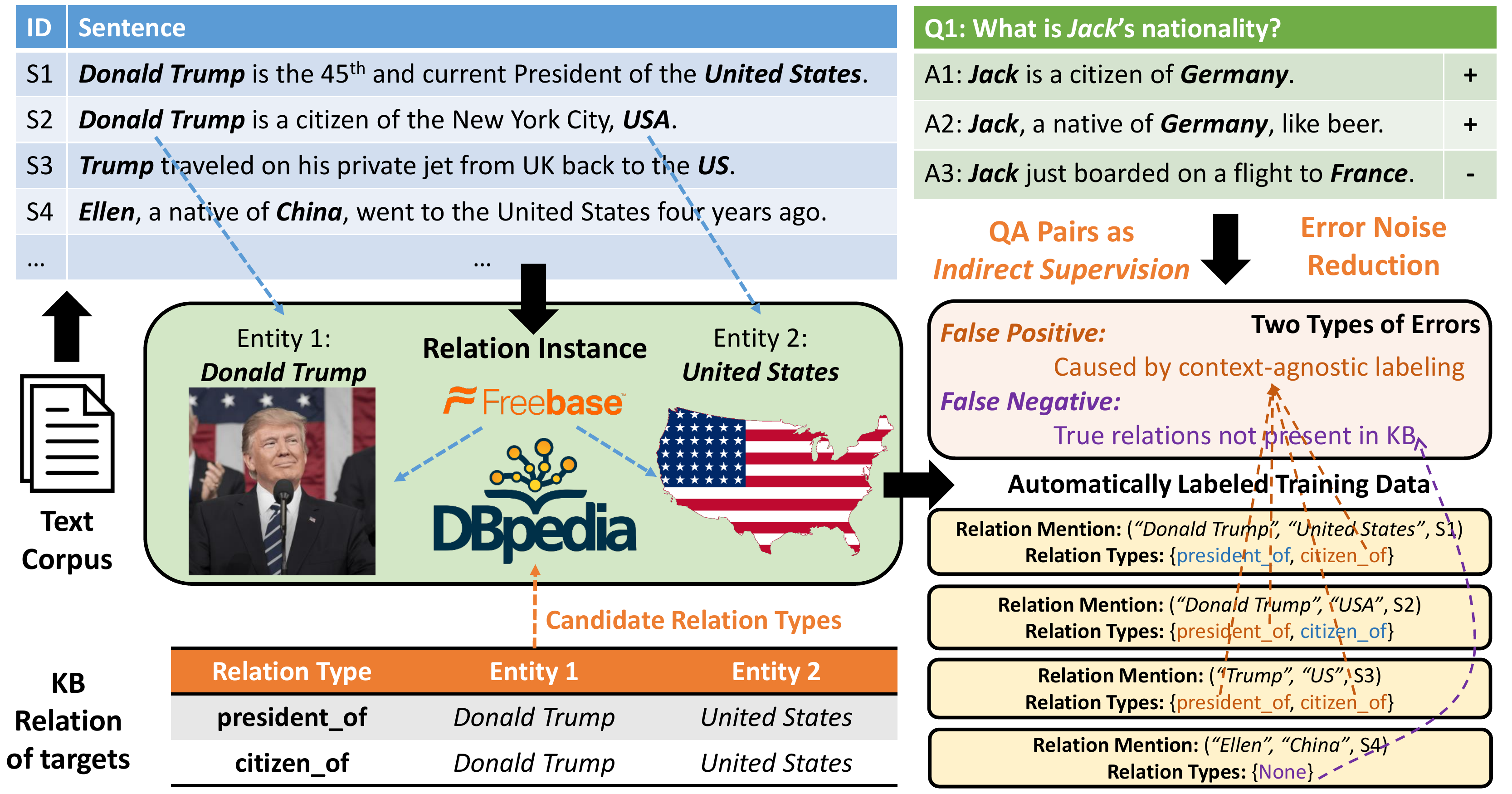}
		\caption{Distant supervision generates training data by linking relation mentions in sentences S1-S4 to KB and assigning the linkable relation types to all relation mentions. Those unlinkable entity mention pairs are treated as negative examples. This automatic labeling process may cause errors of false positives (highlighted in red) and false negatives (highlighted in purple). QA pairs provide indirect supervision for correcting such errors.}
		\label{figure:motivated_example}
    \vspace{0.3cm}
	\end{small}
\end{figure}

\section{Introduction}
Relation extraction is an important task for understanding massive text corpora by turning unstructured text data into relation triples for further analysis.
For example, it detects the relationship ``\texttt{president\_of}'' between entities ``\textit{Donald Trump}'' and ``\textit{United States}'' in a sentence. Such extracted information can be used for more downstream text analysis tasks (e.g. serving as primitives for information extraction and knowledge base (KB) completion, and assisting question answering systems).

\begin{figure*}[th]
	\centering
	\begin{small}
		\includegraphics[width=\textwidth]{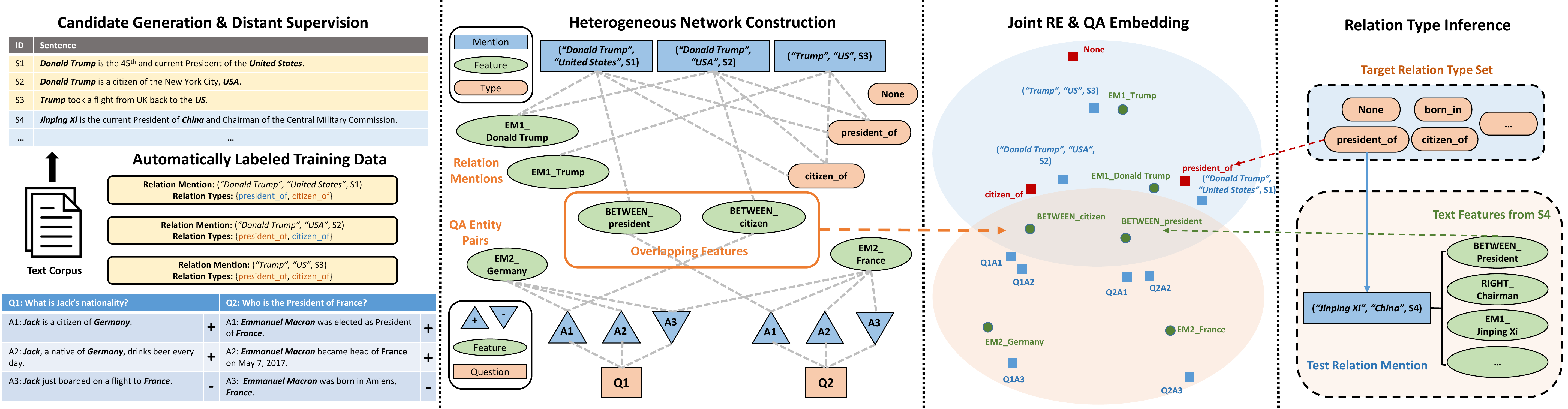}
		\caption{Overall Framework.}
		\label{fig:framework}
	\end{small}
	\vspace{-0.4cm}
\end{figure*}

Typically, RE systems rely on training data, primarily acquired via human annotation, to achieve satisfactory performance. However, such manual labeling process can be costly and non-scalable when adapting to other domains (e.g. biomedical domain). In addition, when the number of types of interest becomes large, the generation of handcrafted training data can be error-prone. To alleviate such an exhaustive process, the recent trend has deviated towards the adoption of distant supervision (DS). DS replaces the manual training data generation with a pipeline that automatically links texts to a knowledge base (KB). The pipeline has the following steps: (1) detect entity mentions in text; (2) map detected entity mentions to entities in KB; (3) assign, to the candidate type set of each entity mention pair, all KB relation types between their KB-mapped entities. However, the noise introduced to the automatically generated training data is not negligible. There are two major causes of error: incomplete KB and context-agnostic labeling process. If we treat unlinkable entity pairs as the pool of negative examples, false negatives can be commonly encountered as a result of the insufficiency of facts in KBs, where many true entity or relation mentions fail to be linked to KBs (see example in ~\figref{figure:motivated_example}). In this way, models counting on extensive negative instances may suffer from such misleading training data. On the other hand, context-agnostic labeling can engender false positive examples, due to the inaccuracy of the DS assumption that if a sentence contains any two entities holding a relation in the KB, the sentence must be expressing such relation between them. For example, entities ``\textit{Donald Trump}'' and ``\textit{United States}'' in the sentence ``\textit{Donald Trump flew back to United States}'' can be labeled as ``\texttt{president\_of}'' as well as ``\texttt{born\_in}'', although only an out-of-interest relation type ``\texttt{travel\_to}'' is expressed explicitly (as shown in ~\figref{figure:motivated_example}).

\begin{figure}[t]
	\centering
	\begin{small}
		\includegraphics[width=\columnwidth]{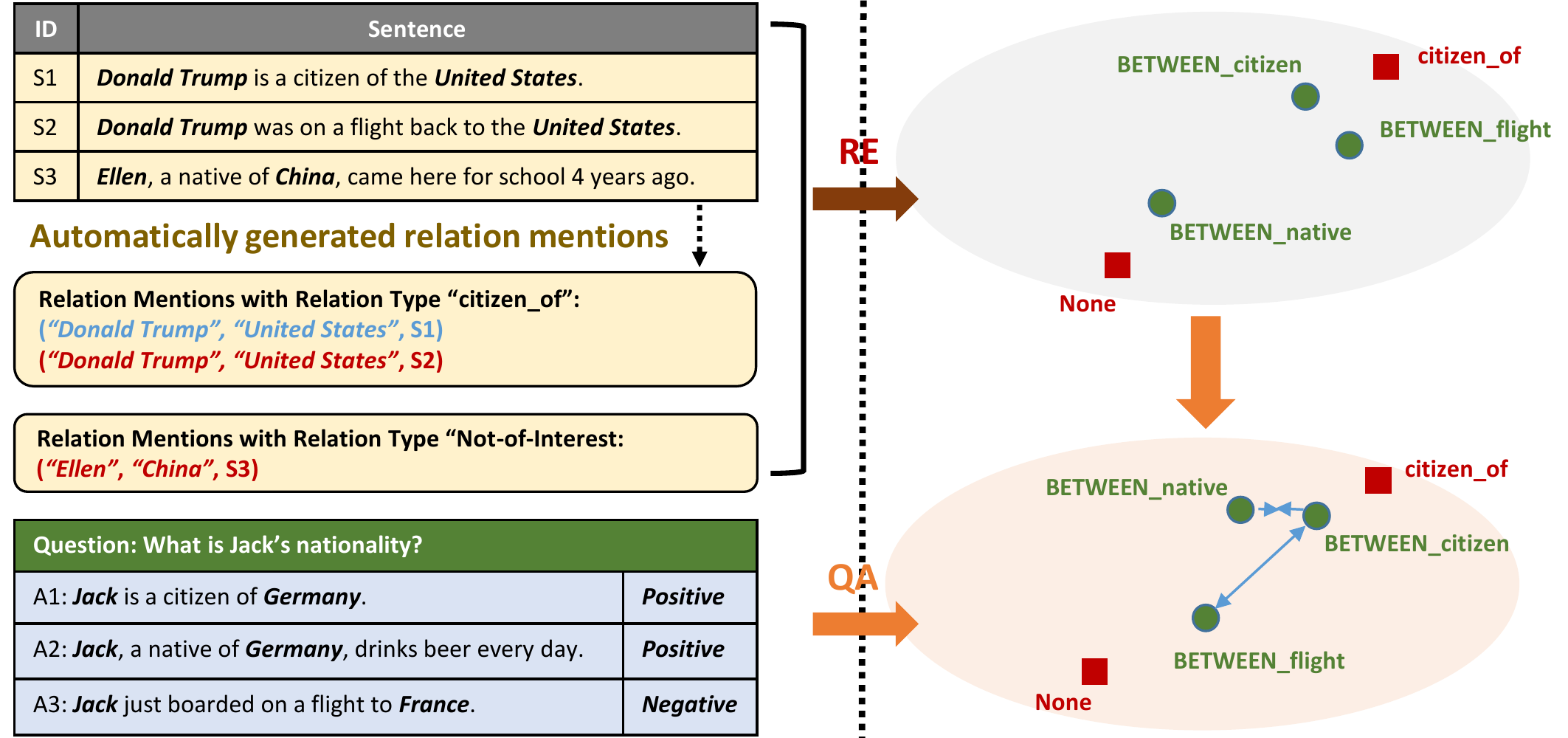}
		\caption{Due to the noise in the automatically generated RE training corpus, the associations between learned feature embeddings and relation types can be affected by the wrongly labeled training examples. However, the idea of QA pairwise interactions has the potential to correct such embedding deviations by bringing extra semantic clues from overlapped features in QA corpus.}
		\label{fig:qa}
	\end{small}
\end{figure}

Towards the goal of diminishing the negative effects by noisy DS training data, distantly supervised RE models that deal with training noise, as well as methods that directly improve the automatic training data generation process have been proposed. These methods mostly involve designing distinct assumptions to remove redundant training information~\cite{Mintz2009DistantSF,Riedel2010ModelingRA,Hoffmann2011KnowledgeBasedWS,Lin2016NeuralRE}. For example, method applied in~\cite{Riedel2010ModelingRA,Hoffmann2011KnowledgeBasedWS} assumes that for each relation triple in the KB, at least one sentence might express the relation instead of all sentences. Moreover, these noise reduction systems usually only address one type of error, either false positives or false negatives. Hence, current methods handling DS noises still have the following challenges:
\begin{enumerate}[leftmargin=12pt]\itemsep+0.0cm
	\item Lack of trustworthy sources: Current de-noising methods mainly focus on recognizing labeling mistakes from the labeled data itself, assisted by pre-defined assumptions or patterns. They do not have external trustworthy sources as guidance to uncover incorrectly labeled data, while not at the expense of excessive human efforts. Without other separate information sources, the reliability of false label identification can be limited.
	\item Incomplete noise handling: Although both false negative and false positive errors are observed to be significant, most existing works only address one of them.
\end{enumerate}
In this paper, to overcome the above two issues derived from relation extraction with distant supervision, we study the problem of relation extraction with indirect supervision from external sources. Recently, the rapid emergence of QA systems promotes the availability of user feedback or datasets of various QA tasks. We investigate to leverage QA, a downstream application of relation extraction, to provide additional signals for learning RE models. Specifically, we use datasets for the task of answer sentence selection to facilitate relation typing. Given a domain-specific corpus and a set of target relation types from a KB, we aim to detect relation mentions from text and categorize each in context by target types or Non-Target-Type (\texttt{None}) by leveraging an independent dataset of QA pairs in the same domain. We address the above two challenges as follows: (1) We integrate indirect supervision from another same-domain data source in the format of QA sentence pairs, that is, each question sentence maps to several positive (where a true answer can be found) and negative (where no answer exists) answer sentences. We adopt the principle that for the same question, positive pairs of (question, answer) should be semantically similar while they should be dissimilar from negative pairs. (2) Instead of differentiating types of labeling errors at the instance level, we concentrate on how to better learn semantic representation of features. Wrongly labeled training examples essentially misguide the understanding of features. It increases the risk of having a non-representative feature learned to be close to a relation type and vice versa. Therefore, if the feature learning process is improved, potentially both types of error can be reduced. (See how QA pairs improve the feature embedding learning process in ~\figref{fig:qa}).

To integrate all the above elements, a novel framework, \textsc{ReQuest}, is proposed. First, \textsc{ReQuest} constructs a heterogeneous graph to represent three kinds of objects: relation mentions, text features and relation types for RE training data labeled by KB linking. Then, \textsc{ReQuest} constructs a second heterogeneous graph to represent entity mention pairs (include question, answer entity mention pairs) and features for QA dataset. These two graphs are combined into a single graph by overlapped features. We formulate a global objective to jointly embed the graph into a low-dimensional space where, in that space, RE objects whose types are semantically close also have similar representations and QA objects linked by positive (question, answer) entity mention pairs of a same question should have close representations. In particular, we design a novel margin-based loss to model the semantic similarity between QA pairs and transmit such information into feature and relation type representations via shared features. With the learned embeddings, we can efficiently estimate the types for test relation mentions.
In summary, this paper makes the following contributions:
\begin{enumerate}[leftmargin=12pt]\itemsep+0.0cm
\item We propose the novel idea of applying indirect supervision from question answering datasets to help eliminate noise from distant supervision for the task of relation extraction.
\item We design a novel joint optimization framework, \textsc{ReQuest}, to extract typed relations in domain-specific corpora.
\item Experiments with two public RE datasets combined with TREC QA demonstrate that \textsc{ReQuest} improves the performance of state-of-the-art RE systems significantly.
\end{enumerate}

\section{Definitions and Problem}
\label{sec:problem}

Our proposed \textsc{ReQuest} framework takes the following input: an automatically labeled training corpus $\D_L$ obtained by linking a text corpus $\D$ to a KB (e.g. Freebase) $\Psi$, a target relation type set $\R$ and a set of QA sentence pairs $\D_{QAS}$ with extract answers labeled.

\smallskip\noindent
\textbf{Entity and Relation Mention.}
An \textit{entity mention} (denoted by $m$) is a token span in text which represents an entity $e$. A \textit{relation instance} $r(e_1,e_2, \ldots, e_n)$ denotes some type of relation $r\in\R$ between multiple entities. In this paper, we focus on binary relations, \ie, $r(e_1,e_2)$.
We define a \textit{relation mention} (denoted by $z$) for some relation instance $r(e_1,e_2)$ as a (ordered) pair of entities mentions of $e_1$ and $e_2$ in a sentence $s$, and represent a relation mention with entity mentions $m_1$ and $m_2$ in sentence $s$ as $z=(m_1, m_2, s)$.

\smallskip\noindent
\textbf{Knowledge Bases and Target Types.}
A KB contains a set of entities $\E_\Psi$, entity types $\Y$ and relation types $\R$, as well as human-curated facts on both relation instances $\I_\Psi = \{r(e_1,e_2)\} \subset \R_\Psi \times \E_\Psi\times
\E_\Psi$, and entity-type facts $\T_\Psi=\{(e,y)\}\subset \E_\Psi\times\Y_\Psi$. \textit{Target relation type set} $\R$ covers a subset of relation types that the users are interested in from $\Psi$, \ie, $\R\subset\R_\Psi$.

\smallskip\noindent
\textbf{Automatically Labeled Training Corpora.}
Distant supervision maps the set of entity mentions extracted from the text corpus to KB entities $\E_\Psi$ with an entity disambiguation system~\cite{mendes2011dbpedia,hoffart2011robust}. Between any two linkable entity mentions $m_1$ and $m_2$ in a sentence, a relation mention $z_i$ is formed if there exists one or more KB relations between their KB-mapped entities $e_1$ and $e_2$. Relations between $e_1$ and $e_2$ in KB are then associated to $z_i$ to form its candidate relation type set $\R_i$, \ie, $\R_i=\{r~|~r(e_1, e_2)\in\R_\Psi\}$.

Let $\Z=\{z_i\}_{i=1}^{N_Z}$ denote the set of extracted relation mentions that can be mapped to KB. Formally, we represent the automatically labeled training corpus $\D_L$ for relation extraction, using a set of tuples $\D_L = \{(z_i, \R_i)\}_{i=1}^{N_Z}$. There exists publicly available automatically labeled corpora such as the NYT dataset~\cite{Riedel2010ModelingRA} where relation mentions have already been extracted and mapped to KB.

\smallskip\noindent
\textbf{QA Entity Mention Pairs.}
The set of QA sentence pairs $\D_{QAS}$ consists of questions $\Q$ in the same domain as the training text corpus. For each question $q_i$, there will be a number of positive sentences $\A_{i}^{+}$, each of which contains a correct answer to the question and another set of negative sentences $\A_{i}^{-}$ where no answer can be found. And the tokens spans of the exact answer in each positive is marked as well. For each question, we extract positive QA (ordered) entity mention pairs $\P_{i}^{+}$ from $\A_{i}^{+}$ and negative entity mention pairs $\P_{i}^{-}$ from $\A_{i}^{-}$. A positive QA entity mention pair $p_k$ contains an entity mention being asked about (question entity mention $m_1$) and an entity mention serving as the answer (answer entity mention $m_2$) to a question. That being said, we can get one positive QA entity mention pair from each positive answer sentence if both entity mentions can be found. In contrast, A negative QA entity mention pair does not follow such pattern for the corresponding question.

Let \begin{small}$\Q=\{q_i\}_{i=1}^{N_q}$\end{small} denote the set of questions; \begin{small}$\P=\{p_k\}_{k=1}^{N_p}$\end{small} denote all QA entity mention pairs; \begin{small}$\P_{i}^{+}=\{p_{k^+}\}_{{k^+}=1}^{N_i^{+}}$\end{small} denote the set of positive QA entity mention pairs for $q_i$; \begin{small}$\P_{i}^{-}=\{p_{k^-}\}_{{k^-}=1}^{N_i^{-}}$\end{small} denote the set of negative QA entity mention pairs for $q_i$. Formally, the QA entity mention pairs corpus is represented as \begin{small}$\D_{QA} = \{(q_i, \P_{i}^{+}, \P_{i}^{-})\}_{i=1}^{N_q}$\end{small}.
\begin{definition}[Problem Definition]
\textbf{Given} an automatically generated training corpus $\D_L$, a target relation type set $\R\subset\R_\Psi$ and a set of QA sentence pairs $\D_{QAS}$ in the same domain, the relation extraction task \textbf{aims to} (1) extract QA entity mention pairs to generate $\D_{QA}$; (2) estimate a relation type $r^*\in\R\cup$\{\texttt{None}\} for each test relation mention, using both the training corpus and the extracted QA pairs with their contexts.
\end{definition}

\section{Approach}
\label{sec:approach}
\noindent
\textbf{Framework Overview.}
We propose an \textit{embedding-based} framework with indirect supervision (illustrated in ~\figref{fig:framework}) as follows:

\begin{enumerate}[leftmargin=12pt]\itemsep+0.0cm
\item Generate text features for each relation mention or QA entity mention pair, and construct a heterogeneous graph using four kinds of objects in combined corpus, namely relation mentions from RE corpus, entity mention pairs from QA corpus, target relation types and text features to encode aforementioned signals in a unified form (\secref{subsec:network_construction}).

\item Jointly embed relation mentions, QA pairs, text features, and type labels into two low-dimensional spaces connected by shared features, where close objects tend to share the same types or questions (\secref{subsec:model}).

\item Estimate type labels $r^*$ for each test relation mention $z$ from learned embeddings, by searching the target type set $\R$ (\secref{subsec:algorithm}).
\end{enumerate}

\subsection{Heterogeneous Network Construction}
\label{subsec:network_construction}
\noindent
\textbf{Relation Mentions and Types Generation.}
We get the relation mentions along with their heuristically obtained relation types from the automatically labeled training corpus $\D_L$. And we randomly sample a set of unlinkable entity mention pairs as the negative relation mentions (\ie, relation mentions assigned with type ``\texttt{None}'').

\smallskip
\noindent
\textbf{QA Entity Mention Pairs Generation.}
We apply Stanford NER~\cite{manning2014stanford} to extract entity mentions in each question or answer sentence. First, we detect the target entity being asked about in each question sentence. For example, in the question ``\textit{Who is the president of United States}'', the question entity is ``\textit{United States}''. In most cases, a question only contains one entity mention and for those containing multiple entity mentions, we notice the question entity is mostly mentioned at the very last. Thus, we follow this heuristic rule to assign the lastly occurred entity mention to be the question entity mention $m_0$ in each question sentence $q_i$. Then, in each positive answer sentence of $q_i$, we extract the entity mention with matched head token and smallest edit string distance to be the question entity mention $m_1$, and the entity mention matching the exact answer string to be the answer entity mention $m_2$. Then we form a positive QA entity mention pair with its context $s$, $p_k = (m_1, m_2, s) \in \P_{i}^{+}$ for $q_i$. If either $m_1$ or $m_2$ can not be found, this positive answer sentence is dropped. We randomly select pairs of entity mentions in each negative answer sentence to be negative QA entity mention pairs for $q_i$ (\eg, if a negative sentence includes 3 entity mentions, we randomly select negative examples from the $3 \cdot 2 \cdot 1 = 6$ different pairs of entity mentions in total, if we ignore the order), with each negative example marked as $p_k\prime = (m_1\prime, m_2\prime, s\prime) \in \P_{i}^{-}$ for $q_i$.

\smallskip
\noindent
\textbf{Text Feature Extraction.}
We extract lexical features of various types from not only the mention itself (\eg, head token), as well as the context $s$ (\eg, bigram) in a POS-tagged corpus.
It is to capture the syntactic and semantic information for any given relation mentions or entity mention pairs.
See~\tabref{tbl:features} for all types of text features used, following those in~\cite{Mintz2009DistantSF,chan2010exploiting} (excluding the dependency parse-based features and entity type features).

We denote the set of $M_z$ unique features extracted from relation mentions $\Z$ as \begin{small}$\F_{z}=\{f_j\}_{j=1}^{M_z}$\end{small} and the set of $M_{QA}$ unique features extracted of QA entity mention pairs $\P$ as \begin{small}$\F_{QA}=\{f_j\}_{j=1}^{M_{QA}}$\end{small}. As our embedding learning process will combine these two sets of features and their shared ones will act as the bridge of two embedding spaces, we denote the overall feature set as \begin{small}$\F=\{f_j\}_{j=1}^{M}$\end{small}.

\smallskip
\noindent
\textbf{Heterogeneous Network Construction.}
After the nodes generation process, we construct a heterogeneous network connected by text features, relation mentions, relation types, questions, QA entity mention pairs, as shown in the second column of ~\figref{fig:framework}.

\begin{table*}[t]
\begin{center}
	\begin{tabularx}{0.94\linewidth}{lll}
		\hline
		\textbf{Feature} & \textbf{Description} & \textbf{Example} \\
		\hline
		Entity mention (EM) head & Syntactic head token of each entity mention & ``\textit{HEAD\_EM1\_Trump}''
		\\
		Entity Mention Token & Tokens in each entity mention & ``\textit{TKN\_EM1\_Donald}''
		\\
		Tokens between two EMs & Each token between two EMs & ``\textit{is}'', ``\textit{the}'', ``\textit{current}'', ``\textit{President}'', ``\textit{of}'', ``\textit{the}''
		\\
		Part-of-speech (POS) tag & POS tags of tokens between two EMs & ``\textit{VBZ}'', ``\textit{DT}'', ``\textit{JJ}'', ``\textit{NN}'', ``\textit{IN}'', ``\textit{DT}'' \\ 
		Collocations & Bigrams in left/right 3-word window of each EM & ``\textit{NYC native}'', ``\textit{native Donald}'', ...
		\\ 
		Entity mention order & Whether EM 1 is before EM 2 & ``\textit{EM1\_BEFORE\_EM2}''\\ 
		Entity mention distance & Number of tokens between the two EMs & ``\textit{EM\_DISTANCE\_6}'' \\
		Entity mention context & Unigrams before and after each EM & ``\textit{native}'', ``\textit{is}'', ``\textit{the}'', ``\textit{.}'' \\
		Special pattern & Occurrence of pattern ``em1\_in\_em2'' & ``\textit{PATTERN\_NULL}''
		\\
		Brown cluster (learned on $\D$) & Brown cluster ID for each token  & ``\textit{8\_1101111}'',~``\textit{12\_111011111111}'' \\ 
		\hline
	\end{tabularx}
\caption{Text features for relation mentions used in this work~\cite{Zhou2005ExploringVK,Riedel2010ModelingRA} (excluding dependency parse-based features and entity type features). (``\textit{Donald Trump}'', ``\textit{United States}'') is used as an example relation mention from the sentence ``\textit{NYC native \textbf{Donald Trump} is the current President of the \textbf{United States}.}''.}
\label{tbl:features}
\end{center}
\end{table*}

\subsection{Joint RE and QA Embedding}
\label{subsec:model}
This section first introduces how we model different types of interactions between linkable relation mentions \begin{small}$\Z$\end{small}, QA entity mention pairs \begin{small}$\P$\end{small}, relation type labels \begin{small}$\R$\end{small} and text features \begin{small}$\F$\end{small} into a $d$-dimensional \textit{relation vector space} and a $d$-dimensional \textit{QA pair vector space}. In the relation vector space, objects whose types are close to each other should have similar representation and in the QA pair vector space, positive QA mention pairs who share the same question are close to each other. (\eg, see the 3rd col. in ~\figref{fig:framework}). We then combine multiple objectives and formulate a joint optimization problem.

We propose a novel global objective, which employs a margin-based rank loss~\cite{nguyen2008classification} to model \textit{noisy mention-type associations} and utilizes the second-order proximity idea~\cite{tang2015line} to model \textit{mention-feature (QA pair-feature) co-occurrences}. In particular, we adopt a pairwise margin loss, following the intuition of pairwise rank ~\cite{Rao2016NoiseContrastiveEF} to capture the \textit{interactions between QA pairs}, and the shared features $\F_z \cap \F_{QA}$ between relation mentions $\Z$ and QA pairs $\P$ connect the two vector spaces.

\smallskip
\noindent
\textbf{Modeling Types of Relation Mentions.}
We introduce the concepts of both \textit{mention-feature co-occurrences} and \textit{mention-type associations} in the modeling of relation types for relation mentions in set $Z$.

The first hypothesis involved in modeling types of relation mentions is as follows.

\vspace{-0.1cm}
\begin{hypothesis}[Mention-Feature Co-occurrence]
\label{hypo:co_occurrence}
If two relation mentions share many text features, they tend to share similar types (close to each other in the embedding space). If two features co-occur with a similar set of relation mentions, they tend to have similar embedding vectors.
\end{hypothesis}

This is based on the intuition that if two relation mentions share many text features, they have high distributional similarity over the set of text features $\F_z$ and likely they have similar relation types. On the other hand, if text features co-occur with many relation mentions in the corpus, such features tend to represent close type semantics. For example, in sentences $s_1$ and $s_4$ in the first column of ~\figref{fig:framework}, the two relation mentions (``\textit{Donald Trump}'', ``\textit{United States}'', $s_1$) and (``\textit{Jinping Xi}'', ``\textit{China}'', $s_4$) share many text features including ``\textit{BETWEEN\_President}'' and they indeed have the same relation type ``\texttt{president\_of}''

Formally, let vectors $\bz_i,~\bc_j\in\RR^d$ represent relation mention $z_i\in\Z$ and text feature $f_j\in\F_z$ in the $d$-dimensional \textit{relation embedding space}. Similar to the distributional hypothesis~\cite{mikolov2013distributed} in text corpora, we apply second-order proximity~\cite{tang2015line} to model the idea in Hypothesis 1 as follows.
\begin{align}
\label{eq:mention_feature_obj}
\L_{ZF} = -\sum_{z_i\in\Z}\sum_{f_j\in\F_z}w_{ij}\cdot\log~p(f_j|z_i),
\end{align}
where \begin{small}$p(f_j|z_i) = \e(\bz_i^T\bc_j)\big/\sum_{f^{\prime}\in\F_z}\e(\bz_{i}^T\bc_{j'})$\end{small} denotes the probability of \begin{small}$f_j$\end{small} generated by \begin{small}$z_i$\end{small}, and \begin{small}$w_{ij}$\end{small} is the co-occurrence frequency between \begin{small}$(z_i, f_j)$\end{small} in corpus \begin{small}$\D$\end{small}.

For the goal of efficient optimization, we apply negative sampling strategy~\cite{mikolov2013distributed} to sample multiple \textit{false} features for each \begin{small}$(z_i,f_j)$\end{small} based on some \textit{noise distribution} \begin{small}$P_n(f)\propto D_f^{3/4}$\end{small}~\cite{mikolov2013distributed} (with \begin{small}$D_f$\end{small} denotes the number of relation mentions co-occurring with $f$). Term \begin{small}$\log~p(f_j|z_i)$\end{small} in Eq.~\eqref{eq:mention_feature_obj} is replaced with the term as follows.
\begin{align}
\label{eq:mention_feature_obj_neg}
\log~\sigma(\bz_i^T\bc_j) + \sum_{v=1}^V\EE_{f_{j'}\sim P_n(f)}\big[\log~\sigma(-\bz_i^T\bc_{j'})\big],
\end{align}
where \begin{small}$\sigma(x)=1/\big(1+\exp(-x)\big)$\end{small} is the sigmoid function. The first term in Eq.~\eqref{eq:mention_feature_obj_neg} models the observed co-occurrence, and the second term models the \begin{small}$V$\end{small} negative feature samples.

In \begin{small}$D_L$\end{small}, each relation mention \begin{small}$z_i$\end{small} is associated with a set of candidate types \begin{small}$\R_i$\end{small} in a context-agnostic setting, which leads to some false associations between \begin{small}$z_i$\end{small} and \begin{small}$r\in\R_i$\end{small} (\ie, false positives). For example, in the first column of ~\figref{fig:framework}, the two relation mentions (``\textit{Donald Trump}'', ``\textit{United States}'', $s_1$) and (``\textit{Donald Trump}'', ``\textit{USA}'', $s_2$) are assigned to the same relation types while each mention actually only has one true type. To handle such conflicts, we use the following hypothesis to model the associations between each linkable relation mention \begin{small}$z_i$\end{small} (in set \begin{small}$\Z$\end{small}) and its noisy candidate relation type set \begin{small}$\R_i$\end{small}.

\vspace{-0.05cm}
\begin{hypothesis}[Partial-Label Association]
\label{hypo:partial_label}
A relation mention's embedding vector should be more similar (closer in the low-dimensional space) to its ``most relevant'' candidate type, than to any other non-candidate type.
\end{hypothesis}

Let vector \begin{small}$\br_k\in\RR^d$\end{small} denote relation type \begin{small}$r_k\in\R$\end{small} in the embedding space, the similarity between \begin{small}$(z_i,r_k)$\end{small} is defined as the dot product of their embedding vectors, \ie, \begin{small}$\phi(z_i,r_k)=\bz_i^T\br_k$\end{small}. \begin{small}$\Rn_i=\R\setminus\R_i$\end{small} denotes the set of \textit{non-candidate types}. We extend the margin-based loss in~\cite{nguyen2008classification} to define a partial-label loss \begin{small}$\ell_i$\end{small} for each linkable relation mention \begin{small}$z_i\in\M_L$\end{small} as follows.
\begin{align}
\label{eq:partial_label_loss}
\ell_{i} = \max\Big\{0, 1 - \Big[\max_{r\in\R_i}\phi(z_i, r)  - \max_{r'\in \Rn_i}\phi(z_i, r')\Big]\Big\}.
\end{align}

To comprehensively model the types of relation mentions, we integrate the modeling of mention-feature co-occurrences and mention-type associations by the following objective, so that feature embeddings also participate in modeling the relation type embeddings.
\begin{align}
\label{eq:relation_mention_obj}
O_{Z} = \L_{ZF} + \sum_{i=1}^{N_Z} \ell_i + \frac{\lambda}{2} \sum_{i=1}^{N_Z}\|\bz_i\|_2^2 + \frac{\lambda}{2} \sum_{k=1}^{K_r} \|\br_k\|_2^2,
\end{align}
where tuning parameter \begin{small}$\lambda > 0$\end{small} on the regularization terms is used to control the scale of the embedding vectors.

\smallskip
\noindent
\textbf{Modeling Associations between QA Entity Mention Pairs.}
We follow Hypothesis 1 to model the QA pair-feature co-occurrence in a similar way. Formally, let vectors \begin{small}$\bp_i,\bc'_j\in\RR^d$\end{small} represent QA entity mention pair \begin{small}$p_i\in\P$\end{small} and text features (for entity mentions) \begin{small}$f_j\in\F_{QA}$\end{small}in a $d$-dimensional \textit{QA entity pair embedding space}, respectively. We model the corpus-level co-occurrences between QA entity mention pairs and text features by second-order proximity as follows.
\begin{align}
\label{eq:entity_mention_feature_obj_neg}
\L_{PF} = -\sum_{p_i\in\P}\sum_{f_j\in\F_{QA}}w_{ij}\cdot\log~p(f_j|p_i),
\end{align}
where the term \begin{small}$\log~p(f_j|p_i)$\end{small} is defined as \begin{small}$\log~p(f_j|p_i) = \log~\sigma(\bp_i^T\bc'_j) + \sum_{v=1}^V\EE_{f_{j'}\sim P_n(f)}\big[\log~\sigma(-\bp_i^T\bc'_{j'})\big]$\end{small}.

For each QA entity mention pair, if we consider it as a relation mention with an unknown type, intuitively, positive pairs sharing a same question are relation mentions with the same relation type or more specifically, are semantically similar relation mentions. In contrast, a positive pair and a negative pair for a question should be semantically far away from each other. For example, in ~\figref{fig:qa}, the embeddings of the entity mention pair in answer sentence $A_1$ should be close to the pair in $A_2$ while far away from the pair in $A_3$. To impose such idea, we model the interactions between QA entity mention pairs based on the following hypothesis.

\begin{hypothesis}[QA Pairwise Interaction]
\label{hypo:qa_pairs_interaction}
A positive QA entity mention pair's embedding vector should be more similar (closer in the low-dimensional space) to any other positive QA entity mention pair, than to any negative QA entity mention pair of the same question.
\end{hypothesis}

Specifically, we use vector \begin{small}$\bp_k\in\RR^d$\end{small} to represent a positive QA entity mention pair $p_k$ in the embedding space. The similarity between two QA entity mention pairs $p_{k1}$ and $p_{k2}$ is defined as the dot product of their embedding vectors. For a positive QA entity mention pair $p_k$ of a question $q_i$ (e.g. \begin{small}$p_k \in \P_i^+$\end{small}), we define the pairwise margin-based loss as follows.
\begin{align}
\label{eq:entity_mention_feature_obj_neg}
\ell_{i,k} = \sum_{p_{k_1} \in \P_i^+, p_{k_2} \in \P_i^-, k_1 \neq k} \max\Big\{0, 1 - \Big[\phi(p_k, p_{k_1})  - \phi(p_k, p_{k_2})\Big]\Big\}.
\end{align}

To integrate both the modeling of QA pair-feature co-occurrence and QA pairs interaction, we formulate the following objective.
\begin{align}
\label{eq:entity_mention_obj}
O_{QA} = \L_{PF} + \sum_{i=1}^{N_Q} \sum_{k=1}^{N_i^+} \ell_{i,k} + \frac{\lambda}{2} \sum_{k=1}^{N_P}\|\bp_k\|_2^2.
\end{align}
By doing so, we can extend the semantic relationships between QA pairs to feature embeddings, such that features of close QA pairs also have similar embeddings. Thus, the learned embeddings of text features from QA corpus carry semantic information inferred from QA pairs. The shared features can propagate such extra semantic knowledge into relation vector space and help better learn the semantic embeddings of both text features and relation types. While feature embeddings of both false positive or false negative examples in the training corpus can deviate towards unrepresentative relation types, the transmitted knowledge from QA space has the potential to adjust such semantic inconsistency. For example, as illustrated in ~\figref{fig:qa}, the false labeled examples in $s_2$ and $s_3$ lead the features ``\textit{BETWEEN\_flight}'' and ``\textit{BETWEEN\_native}'' to be close to ``\texttt{citizen\_of}'' and ``\texttt{None}'' type respectively. After injecting the QA pairwise interactions from the example question, these wrongly placed features are brought back towards the relation types they actually indicate. Minimizing the objective \begin{small}$O_{QA}$\end{small} yields an QA pair embedding space where, in that space, positive QA mention pairs who share the same question are close to each other.
\begin{algorithm}[t]
	\begin{small}
		\DontPrintSemicolon
		\KwIn{labeled training corpus $\D_L$, text features \begin{small}$\{\F\}$\end{small}, regularization parameter $\lambda$, learning rate $\alpha$, number of negative samples $V$, dim. $d$}
		\KwOut{relation mention/QA entity mention pair embeddings \begin{small}$\{\bz_i\}$/$\{\bp_k\}$\end{small},
			feature embeddings \begin{small}$\{\bc_j\},\{\bc'_j\}$\end{small}, relation type embedding \begin{small}$\{\br_k\}$\end{small}}
		Initialize: vectors \begin{small}$\{\bz_i\}$,$\{\bp_k\}$,$\{\bc_j\}$,$\{\bc'_j\}$,$\{\br_k\}$\end{small} as random vectors\;
		\While{$\O$ in Eq.~\eqref{eq:objective} not converge}{
			Sample one component \begin{small}$O_{cur}$\end{small} from \begin{small}$\{O_Z,~O_{QA}\}$\end{small}

			\If{\begin{small}$O_{cur}$\end{small} is \begin{small}$O_Z$\end{small}}{
				Sample a mention-feature co-occurrence $w_{ij}$; draw $V$ negative samples; update \{$\bz$, $\bc$\} based on $\L_{ZF}$\;
				Sample a relation mention $z_i$; get its candidate types $\R_i$; update $\bz$ and $\{\br\}$ based on \begin{small}$\O_Z-\L_{ZF}$\end{small}
			}
			\If{\begin{small}$O_{cur}$\end{small} is \begin{small}$O_{QA}$\end{small}}{
				Sample a pair-feature co-occurrence $w_{ij}$; draw $V$ negative samples; update \{$\bp$, $\bc'$\} based on $\L_{PF}$\;
				Sample an positive QA entity mention pair $p_k$ of question $q_i$; sample one more positive pair and one negative pair of question $q_i$; update $\bp$ based on \begin{small}$\O_{QA}-\L_{PF}$\end{small}
			}
		}
		\caption{Model Learning of \textsc{ReQuest}}
		\label{algorithm:ReQuest}
	\end{small}
\end{algorithm}

\smallskip
\noindent
\textbf{A Joint Optimization Problem.}
Our goal is to embed all the available information for relation mentions and relation types, QA entity mention pairs and text features into a single d-dimensional embedding space. An intuitive solution is to collectively minimize the two objectives $O_{Z}$ and $O_{QA}$ as the embedding vectors of overlapped text features are shared across relation vector space and QA pair vector space. To achieve the goal, we formulate a joint optimization problem as follows.
\begin{align}
\label{eq:objective}
\min_{\{\bz_i\},\{\bc_j\},\{\br_k\},\{\bp_k\},\{\bc'_j\}} \O = \O_{Z} + \O_{QA}.
\end{align}

When optimizing the global objective $O$, the learning of RE and QA embeddings can be mutually influenced as errors in each component can be constrained and corrected by the other. This mutual enhancement also helps better learn the semantic relations between features and relation types. We apply edge sampling strategy~\cite{tang2015line} with a stochastic sub-gradient descent algorithm~\cite{shalev2011pegasos} to efficiently solve Eq.~\eqref{eq:objective}. In each iteration, we alternatively sample from each of the two objectives \begin{small}$\{O_Z,O_M\}$\end{small} a batch of edges (\eg, \begin{small}$(z_i,f_j)$\end{small}) and their negative samples, and update each embedding vector based on the derivatives. The detailed learning process of \textsc{ReQuest} can be seen in Algorithm~\ref{algorithm:ReQuest}. To prove convergence of this algorithm (to the local minimum), we can adopt the proof procedure in~\cite{shalev2011pegasos}.

\subsection{Type Inference}
\label{subsec:algorithm}
\smallskip
To predict the type for each test relation mention $z$, we search for nearest neighbor in the target relation type set \begin{small}$\R$\end{small}, with the learned embeddings of features and relation types (\ie, \begin{small}$\{\bc_i\}$\end{small}, \begin{small}$\{\bc'_i\}$, $\{\br_k\}$\end{small}). Specifically, we represent test relation mention $z$ in our learned relation embedding space by \begin{small}$\bz = \sum_{f_j\in\F_z(z)}\bc_{j}$\end{small} where \begin{small}$\F_z(z)$\end{small} is the set of text features extracted from $z$'s local context $s$. We categorize $z$ to \texttt{None} type if the similarity score is below a pre-defined threshold (e.g. $\eta>0$).

\section{Experiments}
\label{sec:exp}
\subsection{Data Preparation and Experiment Setting}
\label{sec:data}
Our experiments consists of two different type of datasets, one for relation extraction and another answer sentence selection dataset for indirect supervision.
Two public datasets are used for relation extraction: \textbf{NYT}~\cite{Riedel2010ModelingRA,Hoffmann2011KnowledgeBasedWS}and \textbf{KBP}~\cite{ling2012fine,ellislinguistic}.
The test data are manually annotated with relation types by their respective authors.
Statistics of the datasets are shown in~\tabref{tbl:data_stats}.
Automatically generated training data by distant supervision on these two training corpora have been used in~\cite{Ren2017CoTypeJE,Riedel2010ModelingRA} and is accessible via public links, as well as the test data\footnote{\url{https://github.com/shanzhenren/CoType/tree/master/data/source}}. The automatic data generation process is the same as described in ~\secref{sec:problem} by utilizing DBpedia Spotlight\footnote{\url{http://spotlight.dbpedia.org/}}, a state-of-the-art entity disambiguation tool, and Freebase, a large entity knowledge base. As for QA dataset, we use the answer sentence selection dataset extracted from \textbf{TREC-QA} dataset~\cite{Wang2007WhatIT} used by many researchers~\cite{Wang2015FAQbasedQA,tan2015lstm,Santos2016AttentivePN}. We obtain the compiled version of the dataset from~\cite{Yao2013AnswerEA,Yao2013AutomaticCO}, which can be accessed via publicly available link\footnote{\url{https://github.com/xuchen/jacana/tree/master/tree-edit-data}}. Then, we parse this QA dataset to generate QA entity mention pairs following the steps described in~\secref{subsec:network_construction}. During this procedure, we drop the question or answer sentences where no valid QA entity mention pairs can be found. The statistics of this dataset is presented in ~\tabref{tbl:qa_data_stats}.

\smallskip
\noindent
\textbf{Feature Generation.}
This step is run on both relation extraction dataset and preprocessed QA entity mention pairs and sentences. \tabref{tbl:features} lists the set of text features of both relation mentions and QA entity mention pairs used in our experiments. We use a 6-word window to extract context features for each mention (3 words on the left and the right). We apply the Stanford CoreNLP tool~\cite{manning2014stanford} to get POS tags. Brown clusters are derived for each corpus using public implementation\footnote{\url{https://github.com/percyliang/brown-cluster}}. The same kinds of features are used in all the compared methods in our experiments. As the overlapped features in both RE and QA datasets play an important role in the optimization process, we put the statistics of the shared features in ~\tabref{tbl:feature_stats}.
\begin{table}[t]
	\begin{center}
			\begin{tabularx}{0.65\linewidth}{l ll}
				\hline
				\textbf{Data sets} & \textbf{NYT} & \textbf{KBP} \\
				\hline
				$\#$Relation types & 24  & 19 \\ 
				$\#$Documents & 294,977 & 780,549 \\ 
				$\#$Sentences & 1.18M & 1.51M \\ 
				$\#$Training RMs & 353k & 148k \\ 
				$\#$Text features & 2.6M & 1.3M \\ 
				$\#$Test Sentences & 395 & 289 \\ 
				$\#$Ground-truth RMs & 3,880 & 2,209 \\ 
				\hline
			\end{tabularx}
			\caption{Statistics of relation extraction datasets.}
			\label{tbl:data_stats}
	\end{center}
\end{table}
\begin{table}[t]
	\begin{center}
			\begin{tabularx}{0.96\linewidth}{l ll}
				\hline
				\textbf{Versions of QA dataset} & \textbf{COMPLETE} & \textbf{FILTERED} \\
				\hline
				$\#$Questions & 1.4K  & 186 \\ 
				$\#$Positive Answer Sentences & 6.9K & 969 \\ 
				$\#$Negative Answer Sentences & 49K & 5.5K\\ 
				$\#$Positive entity mention pairs & - & 969 \\ 
				$\#$Negative entity mention pairs & - & 28K \\ 
				\hline
			\end{tabularx}
			\caption{Statistics of the answer sentence selection datasets. The complete version is the raw corpus we obtain from the public link. The filtered version is the input to \textsc{ReQuest} after dropping sentences where no valid QA entity mention pair can be found.}
			\label{tbl:qa_data_stats}
	\end{center}
\end{table}

\begin{table}[t]
\begin{center}
\begin{tabularx}{1\linewidth}{lll}
	\hline
	\textbf{Data sets} & \textbf{NYT} & \textbf{KBP} \\
	\hline
	$\%$ distinct shared features with TREC QA & 10.0\%  & 11.6\% \\ 
	$\%$ occurrences of shared features with TREC QA & 90.1\% & 85.6\% \\ 
	\hline
\end{tabularx}
\caption{Statistics of overlapped features. For example, if we have the following observations in NYT and TREC QA respectively: $(f_1, f_1, f_1, f_2, f_3)$ and $(f_1, f_2, f_4)$, then $\%$ distinct shared features with TREC QA of NYT is $66.7\%$ $(f_1, f_2)$ and $\%$ occurrences of shared features with TREC QA of NYT is $80.0\%$.}
\label{tbl:feature_stats}
\end{center}
\end{table}

\smallskip
\noindent
\textbf{Evaluation Sets.}
The provided train/test split are used in NYT and KBP relation extraction datasets.
The relation mentions in test data have been manually annotated with relation types in the released dataset (see~\tabref{tbl:data_stats} for the data statistics).
A \textit{validation set} is created through randomly sampling 10\% of relation mentions from test data, and the rest are used as \textit{evaluation set}.

\smallskip
\noindent
\textbf{Compared Methods.}
We compare \textsc{ReQuest} with its variants which model parts of the proposed hypotheses. Several state-of-the-art relation extraction methods (\eg, supervised, embedding, neural network) are also implemented (or tested using their published codes):
(1) \textbf{DS+Perceptron}~\cite{ling2012fine}: adopts multi-label learning on automatically labeled training data $\D_L$.
(2) \textbf{DS+Kernel}~\cite{mooney2005subsequence}: applies bag-of-feature kernel~\cite{mooney2005subsequence} to train a SVM classifier using $\D_L$;
(3) \textbf{DS+Logistic}~\cite{Mintz2009DistantSF}: trains a multi-class logistic classifier\footnote{We use liblinear package from \url{https://github.com/cjlin1/liblinear}} on $\D_L$;
(4) \textbf{DeepWalk}~\cite{perozzi2014deepwalk}: embeds mention-feature co-occurrences and mention-type associations as a homogeneous network (with binary edges);
(5) \textbf{LINE}~\cite{tang2015line}:
uses second-order proximity model with edge sampling on a feature-type bipartite graph (where edge weight $w_{jk}$ is the number of relation mentions having feature $f_j$ and type $r_k$);
(6) \textbf{MultiR}~\cite{Hoffmann2011KnowledgeBasedWS}: is a state-of-the-art distant supervision method, which models noisy label in $\D_L$ by multi-instance multi-label learning;
(7) \textbf{FCM}~\cite{gormley2015improved}: adopts neural language model to perform compositional embedding;
(8) \textbf{DS+SDP-LSTM}~\cite{xu2015classifying,xu2016improved}: current state-of-the-art in SemEval 2010 Task 8 relation classification task~\cite{sem}, leverages a multi-channel input along the shortest dependency path between two entities into stacked deep recurrent neural network model. We use $\D_L$ to train the model.
(9) \textbf{DS+LSTM-ER}~\cite{Miwa2016EndtoEndRE}: current state-of-the-art model on ACE2005 and ACE2004 relation classification task~\cite{Doddington2004TheAC,li2014incremental}. It is a multi-layer LSTM-RNN based model that captures both word sequence and dependency tree substructure information. We use $\D_L$ to train the model.
(10) \textbf{CoType-RM}~\cite{Ren2017CoTypeJE}: A distant supervised model which adopts the partial-label loss to handle label noise and train the relation extractor.

Besides the proposed joint optimization model, \textbf{ReQuest-Joint}, we conduct experiments on two other variations to compare the performance (1) \textbf{ReQuest-QA\_RE}: This variation optimizes objective \begin{small}$\O_{QA}$\end{small} first and then uses the learned feature embeddings as the initial state to optimize \begin{small}$\O_{Z}$\end{small}; and (2) \textbf{ReQuest-RE\_QA}: It first optimizes \begin{small}$\O_Z$\end{small}, then optimizes \begin{small}$\O_{QA}$\end{small} to finely tune the learned feature embeddings.

\smallskip
\noindent
\textbf{Parameter Settings.}
In the testing of \textsc{ReQuest} and its variants, we set $\eta=0.35$ and $\lambda=10^{-4}$ and $V=3$ based on validation sets.
We stop further optimization if the relative change of $\O$ in Eq.~\eqref{eq:objective} is smaller than $10^{-4}$. The dimensionality of embeddings $d$ is set to $50$ for all embedding methods. For other parameters, we tune them on validation sets and picked the values which lead to the best performance.

\smallskip
\noindent
\textbf{Evaluation Metrics.}
We adopt standard Precision, Recall and F1 score~\cite{mooney2005subsequence,Bach2007ARO} for measuring the performance of relation extraction task. Note that all our evaluations are \textit{sentence-level} or \textit{mention-level} (\ie, context-dependent), as discussed in~\cite{Hoffmann2011KnowledgeBasedWS}.
\subsection{Experiments and Performance Study}
\label{sec:performance}
\begin{table}[t]
	\begin{center}
		\begin{small}
			\begin{tabularx}{0.98\linewidth}{X|l |l}
				\hline
				\textbf{Relation Mention} & \textbf{ReQuest} & \textbf{CoType-RM} \\
				\hline
				.. traveling to \textit{Amman} \textbf{,} \textit{Jordan} .. & /location/location/contains  & \texttt{None} \\ 
				\hline
				The photograph showed \textbf{Gov.} \textit{Ernie Fletcher} \textbf{of} \textit{Kentucky} .. & /people/person/place\_lived & \texttt{None} \\ 
				\hline
				.. \textbf{as chairman of} the \textit{Securities and Exchange Commission} , \textit{Christopher Cox} .. & /business/person/company & \texttt{None} \\ 
				\hline
			\end{tabularx}
			\caption{Case Study.}
			\label{tbl:case_study}
		\end{small}
	\end{center}
\end{table}

\begin{table*}[th]
	\vspace{0.0cm}
	\begin{center}
		\begin{tabularx}{0.97\textwidth}{  l | aaaa | aaaa}
			\hline
			& \multicolumn{4}{c|}{\textbf{NYT}~\cite{Riedel2010ModelingRA,Hoffmann2011KnowledgeBasedWS}} &  \multicolumn{4}{c}{\textbf{KBP}~\cite{ellislinguistic,ling2012fine}} \\
			\textbf{Method}& \textbf{Prec} & \textbf{Rec}  & \textbf{F1} & \textbf{Time}
			& \textbf{Prec} & \textbf{Rec} & \textbf{F1} & \textbf{Time} \\ \hline
			DS+Perceptron~\cite{ling2012fine}
			& 0.068 & \textbf{0.641} & 0.123 & 15min
			& 0.233 & 0.457 & 0.308 & 7.7min
			\\
			DS+Kernel~\cite{mooney2005subsequence}
			& 0.095 & 0.490 & 0.158 & 56hr
			& 0.108 & 0.239 & 0.149 & 9.8hr
			\\
			DS+Logistic~\cite{Mintz2009DistantSF}
			& 0.258 & 0.393 & 0.311 & 25min
			& 0.296 & 0.387 & 0.335 & 14min
			\\
			DeepWalk~\cite{perozzi2014deepwalk}
			& 0.176 & 0.224 & 0.197 & 1.1hr
			& 0.101 & 0.296 & 0.150 & 27min
			\\
			LINE~\cite{tang2015line}
			& 0.335 & 0.329 & 0.332 & 2.3min
			& 0.360 & 0.257 & 0.299 & 1.5min
			\\
			MultiR~\cite{Hoffmann2011KnowledgeBasedWS}
			& 0.338 & 0.327 & 0.333 & 5.8min
			& 0.325 & 0.278 & 0.301 & 4.1min
			\\
			FCM~\cite{gormley2015improved}
			& \textbf{0.553} & 0.154 & 0.240 & 1.3hr
			& 0.151 & \textbf{0.500} & 0.301 & 25min
			\\
			DS+SDP-LSTM~\cite{xu2015classifying,xu2016improved} & 0.307 & 0.532 & 0.389 & 21hr & 0.249 & 0.300 &  0.272 & 10hr
			\\
			DS+LSTM-ER~\cite{Miwa2016EndtoEndRE} & 0.373 & 0.171 & 0.234 & 49hr & 0.338 & 0.106 & 0.161 & 30hr
			\\
			CoType-RM~\cite{Ren2017CoTypeJE} & 0.467 & 0.380 & 0.419 & 2.6min & 0.342 & 0.339 & 0.340 & 1.5min
			\\
			\hline
			\textsc{ReQuest}-QA\_RE
			& 0.407 & 0.437 & 0.422 & 10.2min
			& \textbf{0.459} & 0.300 & 0.363 & 5.3min
			\\
			\textsc{ReQuest}-RE\_QA
			& 0.435 & 0.419 & 0.427 & 8.0min
			& 0.356 & 0.352 & 0.354 & 13.2min
			\\
			\textsc{ReQuest}-Joint
			& 0.404 & 0.480 & \textbf{0.439} & 4.0min
			& 0.386 & 0.410 & \textbf{0.397} & 5.9min
			\\ \hline
		\end{tabularx}
		\caption{Performance comparison on end-to-end relation extraction (at the highest F1 point) on the two datasets.}
		\label{tbl:relation_extraction}
	\end{center}
\end{table*}

\noindent
\textbf{Performance Comparison with Baselines.}
To test the effectiveness of our proposed framework \textsc{ReQuest}, we compare with other methods on the relation extraction task. The precision, recall, F1 scores as well as the model learning time measured on two datasets are reported in~\tabref{tbl:relation_extraction}. As shown in the table, \textsc{ReQuest} achieves superior F1 score on both datasets compared with other models. Among all these baselines, MultiR and CoType-RM handle noisy training data while the remaining ones assume the training corpus is perfectly labeled. Due to their nature of being cautious towards the noisy training data, both MultiR and CoType-RM reach relatively high results confronting with other models that blindly exploit all heuristically obtained training examples. However, as external reliable information sources are absent and only the noise from multi-label relation mentions (while none or only one assigned label is correct) is tackled in these models, MultiR and CoType-RM underperform \textsc{ReQuest}. Especially from the comparison with CoType-RM, which is also an embedding learning based relation extraction model with the idea of partial-label loss incorporated, we can conclude that the extra semantic inklings provided by the QA corpus do help boost the performance of relation extraction.

\smallskip
\noindent
\textbf{Performance Comparison with Ablations.}
We experiment with two variations of \textsc{ReQuest}, \textsc{ReQuest}-QA\_RE and \textsc{ReQuest}-RE\_QA, in order to validate the idea of joint optimization. As presented in~\tabref{tbl:relation_extraction}, both \textsc{ReQuest}-QA\_RE and \textsc{ReQuest}-RE\_QA outperform most of the baselines, with the indirect supervision from QA corpus. However, their results still fall behind \textsc{ReQuest}'s. Thus, separately training the two components may not capture as much information as jointly optimizing the combined objective. The idea of constraining each component in the joint optimization process proves to be effective in learning embeddings to present semantic meanings of objects (e.g. features, types and mentions).

\subsection{Case Study}
\noindent
\textbf{Example Outputs. }
We have done some interesting investigations regarding the type of prediction errors that can be corrected by the indirection supervision from QA corpus. We have analyzed the prediction results on NYT dataset from CoType-RM and \textsc{ReQuest} and find out the top three target relation types that can be corrected by \textsc{ReQuest} are ``\texttt{contains\_location}'', ``\texttt{work\_for}'', ``\texttt{place\_lived}''.
Both the issues of KB incompleteness and context-agnostic labeling are severe for these relation types. For example, there can be lots of not that well-known suburban areas belonging to a city, a state or a country while not marked in KB. And a person can has lived in tens or even hundreds places for various lengths of period. These are hard to be fully annotated into a KB. Thus, the automatically obtained training corpus may end up containing a large percentage of false negative examples for such relation types. On the other hand, there are abundant entity pairs having both ``\texttt{contains\_location}'' and ``\texttt{capital\_of}'', or both ``\texttt{place\_lived}'' and ``\texttt{born\_in}'' relation types in KB. Naturally, training examples of such entity pairs can be greatly polluted by false positives. In this case, it becomes tough to learn semantic embeddings for relevant features of these relation types. However, we notice there are quite a few answer sentences for relevant questions like ``Where is \textit{XXX} located'', ``Where did \textit{XXX} live'',  ``What company is \textit{XXX} with'' in the QA corpus, which plays an important role in adjusting vectors for features that are supposed to be the indicators for these relation types. \tabref{tbl:case_study} shows some prediction errors from CoType-RM that are fixed in \textsc{ReQuest}.
\label{sec:case}

\smallskip
\noindent
\textbf{Study the effect of QA dataset processing on F1 scores.}
\begin{figure}	\includegraphics[width=0.9\columnwidth]{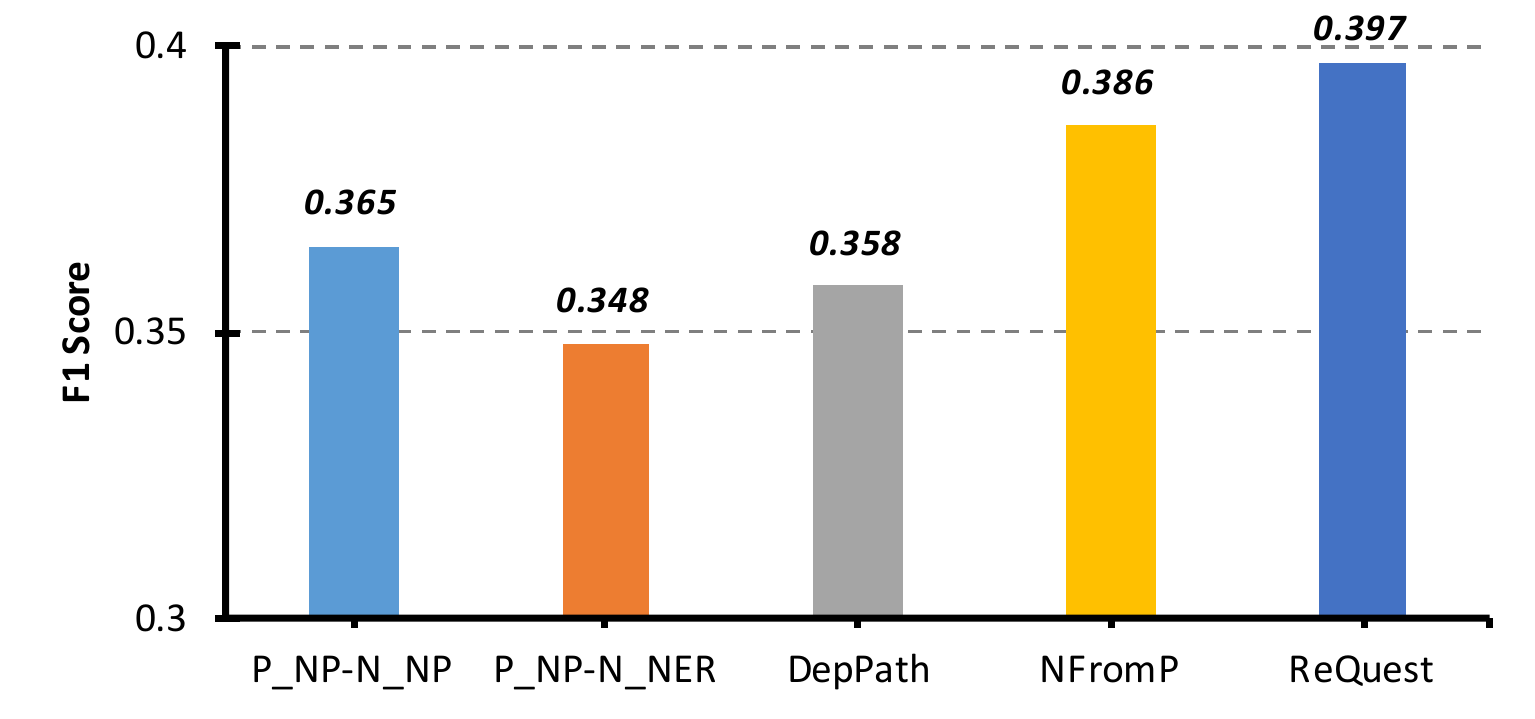}
	\caption{Effect of QA dataset processing on F1 scores. P\_NP-N\_NP: positive QA noun phrase pairs + negative QA noun phrase pairs, P\_NP-N\_NER: positive QA noun phrase pairs + negative QA named entity pairs, DepPath: convert QA sentences to dep paths, NFromP: sample negative QA pairs from both positive and negative answer sentences}
    \vspace{0.5cm}
	\label{tbl:effect}
\end{figure}
As stated in~\secref{subsec:network_construction}, \textsc{ReQuest} uses Stanford NER to extract entity mentions in QA dataset and all QA pairs consist of two entity mentions and if either question or answer entity mention is not found, it drops the sentence. Beyond that, we have conducted experiments with four other ways to construct QA pairs from the raw QA sentences. As shown in \tabref{tbl:qa_data_stats}, we lose many positive QA pairs if we only remain answer (or question) targets that are detected as named entities. Thus, we have tried to keep more positive pairs by relaxing the restriction from named entities to noun phrases. In addition, we have tried to evaluate the performance by 1) keeping negative pairs as named entity pairs or 2) changing them to noun phrase pairs. Besides that, inspired by~\cite{xu2015classifying,xu2016improved}, the third processing variation we have tried is to parse the QA sentences into dependency paths and to extract features from these paths instead of the full sentences. The last one is that, we sample negative QA pairs not only from negative answer sentences, but also from positive sentence when extracting QA pairs. However, \textsc{ReQuest} achieves highest F1 score compared with these four processing variations (as shown in \figref{tbl:effect}) by filtering out all non entity mention answers, keeping full sentences and extracting only positive QA pairs from positive answer sentences.

Although by doing so, \textsc{ReQuest} filters out a large number of question/answer sentences and fewer QA pairs are constructed to provide semantic knowledge for RE, the remaining QA pairs provide cleaner and more consistent information with RE dataset. Thus, it still outperforms the other variations. Another interesting highlight is the comparison between using negative named entity pairs and using negative noun phrase pairs when positive QA pairs are formed by noun phrases. Although enforcing named entities is more consistent with RE datasets, a trade-off exists when the data format of positive and negative QA pairs are inconsistent. As we can see from the bar chart, the performance by using negative noun phrase pairs is better than negative named entity pairs.

\section{Related Work}
\label{sec:related}
Classifying relation types between entities in a certain sentence and automatically extracting them from large corpora plays a key role in information extraction and natural language processing applications and thus has been a hot research topic recently.
Even though many existing knowledge bases are very large, they are still far from complete.
A lot of information is hidden in unstructured data, such as natural language text. Most tasks focus on knowledge base completion (KBP)~\cite{Surdeanu2014OverviewOT} as a goal of relation extraction from corpora like New York Times (NYT)~\cite{Riedel2010ModelingRA}.
Others extract valuable relation information from community question-answer texts, which may be unique to other sources~\cite{Savenkov2015RelationEF}.

For supervised relation extraction, feature-based methods~\cite{sem} and neural network techniques~\cite{socher2011semi,ebrahimi2015chain} are most common. Most of them jointly leverage both semantic and syntactic features~\cite{Miwa2016EndtoEndRE}, while some use multi-channel input information as well as shortest dependency path to narrow down the attention~\cite{xu2015classifying,xu2016improved}.
Two of he aforementioned papers perform the best on the SemEval-2010 Task 8 and constitutes our neural baseline methods.

However, most of these methods require large amount of annotated data, which is time consuming and labor intensive.
To address this issue, most researchers align plain text with knowledge base by \textit{distant supervision}~\cite{Mintz2009DistantSF} for relation extraction.
However, distant supervision inevitably accompanies with the wrong labeling problem.
To alleviate the wrong labeling problem, multi-instance and multi-label learning are used~\cite{Riedel2010ModelingRA,Hoffmann2011KnowledgeBasedWS}. Others~\cite{Ren2017CoTypeJE,li2014incremental} propose joint extraction of typed entities and relations as joint optimization problem and posing cross-constraints of entities and relations on each other. Neural models with selective attention~\cite{Lin2016NeuralRE} are also proposed to automatically reduce labeling noise.

The distant supervision provides one solution to the cost of massive training data.
However, traditional DS methods mostly only exploit one specific kind of
indirect supervision knowledge - the relations/facts in a given knowledge base, thus often suffer from the problem of lack of supervision.
There exist other \textit{indirect supervision} methods for relation extraction, where some utilize globally and cross sentence boundary supervision~\cite{Quirk2016DistantSF,Han2016GlobalDS}, some leverage the power of passage retrieval model for providing relevance feedback on sentences~\cite{Xu2013FillingKB}, and others~\cite{Banko2007OpenIE,Poon2008JointUC,Toutanova2015RepresentingTF}. Recently, with the prevalence of reinforcement learning applications, many information extraction and relation extraction tasks have adopted such techniques to boost existing approaches~\cite{Narasimhan2016ImprovingIE,Kanani2012SelectingAF}.
Our methodology follows the success of indirect supervision, by adding question-answering pairs as another source of supervision for relation extraction task along with knowledge base auto-labeled distant supervision as well as partial supervision.

Another indirect supervision source we use in the paper, passage retrieval, as described here, is the task of retrieving only the portions of a document that are relevant to a particular information need.
It could be useful for limiting the amount of non-relevant material presented to a searcher, or for helping the searcher locate the relevant portions of documents more quickly.
Passage retrieval is also often an intermediate step in other information retrieval tasks, like question answering~\cite{Savenkov2016WhenAK, Ittycheriah2000IBMsSQ,Elworthy2000QuestionAU,Khalid2008PassageRF} and combining with summarization.
Some passage retrieval approaches~\cite{Wade2005PassageRA} include calculating query-likelihood and relevance modeling~\cite{Clarke2000QuestionAB}, others show that language model approaches used for document retrieval can be applied to answer passage retrieval~\cite{CorradaEmmanuel2003AnswerPR}.
Following the success of passage retrieval usage in question-answering pipelines, to the best of our knowledge, we are the first to utilize passage retrieval, or specifically, answer sentence selection from question-answer pairs to provide additional indirect feedback and supervision for relation extraction task.

\section{Conclusion}
We present a novel study on indirect supervision (from question-answering datasets) for the task of relation extraction. We propose a framework, \textsc{ReQuest}, that embeds information from both training data automatically generated by linking to knowledge bases and QA datasets, and captures richer semantic knowledge from both sources via shared text features so that better feature embeddings can be learned to infer relation type for test relation mentions despite the noisy training data. Our experiment results on two datasets demonstrate the effectiveness and robustness of \textsc{ReQuest}. Interesting future work includes identifying most relevant QA pairs for target relation types, generating most effective questions to collect feedback (or answers) via crowd-sourcing, and exploring approaches other than distant supervision~\cite{Riedel2013RelationEW,Artzi2013WeaklySL}.

\begin{acks}
Research was sponsored in part by the U.S. Army Research Lab. under Cooperative Agreement No. W911NF-09-2-0053 (NSCTA), National Science Foundation IIS 16-18481, IIS 17-04532, and IIS-17-41317, and grant 1U54GM114838 awarded by NIGMS through funds provided by the trans-NIH Big Data to Knowledge (BD2K) initiative (www.bd2k.nih.gov). The views and conclusions contained in this paper are those of the authors and should not be interpreted as representing any funding agencies.
\end{acks}

\bibliographystyle{ACM-Reference-Format}
\bibliography{sigproc}

\end{document}